\documentclass[lettersize,journal]{IEEEtran}
\usepackage{caption}
\usepackage{subcaption}
\usepackage{amsmath,amsfonts}
\usepackage{algorithmic}
\usepackage{algorithm}
\usepackage{array}
\usepackage{subcaption}
\expandafter\def\csname ver@subfig.sty\endcsname{}
\usepackage{textcomp}
\usepackage{stfloats}
\usepackage{url}
\usepackage{verbatim}
\usepackage{graphicx}
\usepackage{cite}
\usepackage{hyperref}
\usepackage{multirow}
\usepackage[table]{xcolor}

\begin{document}

\title{Performance and Efficiency of Climate In-Situ Data Reconstruction: Why Optimized IDW Outperforms kriging and Implicit Neural Representation}

\author{Jakub Walczak
\thanks{Manuscript received April 19, 2021; revised August 16, 2021.}
\thanks{Jakub Walczak is affiliated to Lodz university of Technology, Lodz, Poland}
\thanks{\includegraphics[width=0.015\textwidth]{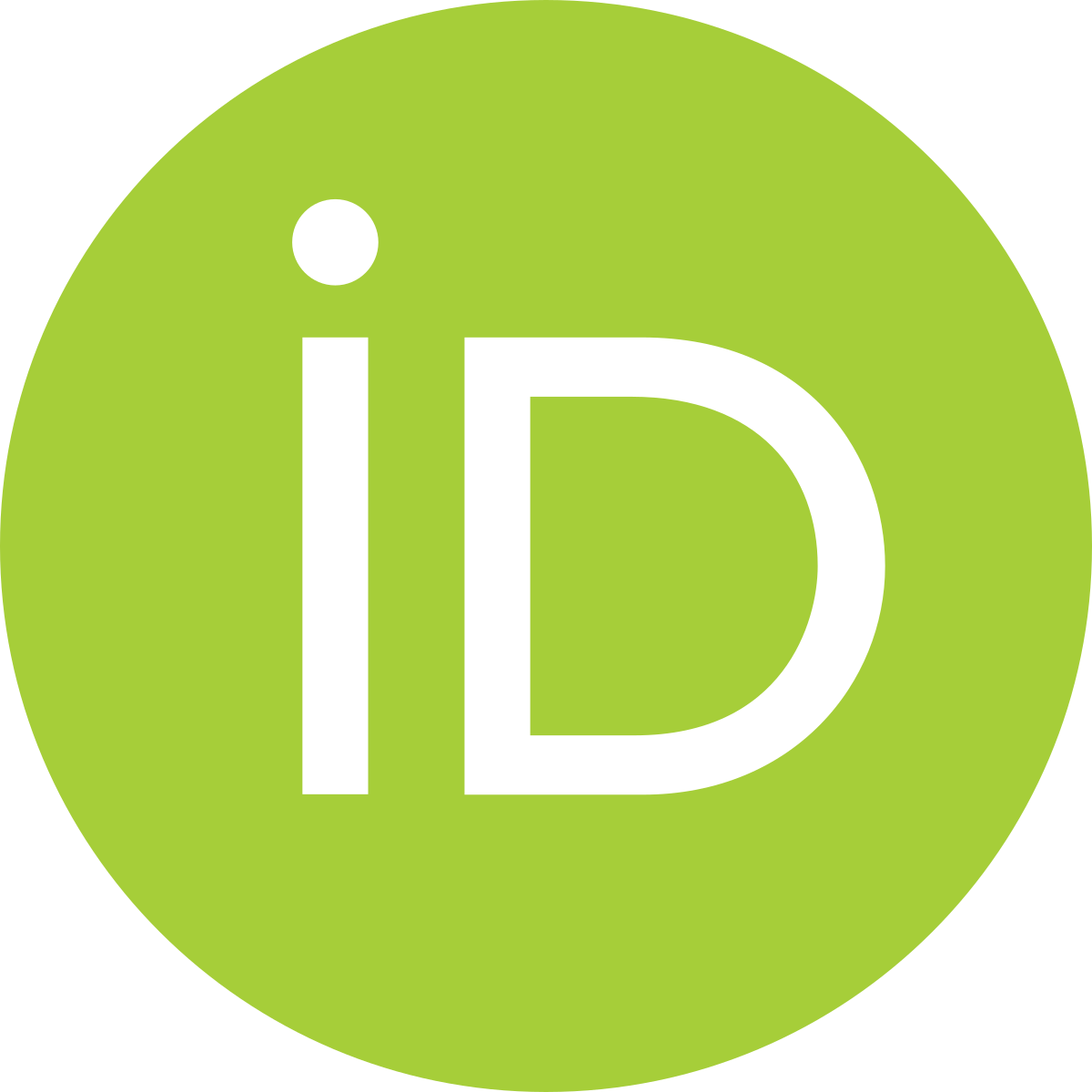} Jakub Walczak: https://orcid.org/0000-0002-5632-9484}
}

\markboth{Journal of \LaTeX\ Class Files,~Vol.~14, No.~8, August~2021}%
{Shell \MakeLowercase{\textit{et al.}}: A Sample Article Using IEEEtran.cls for IEEE Journals}

\IEEEpubid{0000--0000/00\$00.00~\copyright~2021 IEEE}

\maketitle

\begin{abstract}
This study evaluates three reconstruction methods for sparse climate data: the simple inverse distance weighting (IDW), the statistically grounded ordinary kriging (OK), and the advanced implicit neural representation model (MMGN architecture). All methods were optimized through hyper-parameter tuning using validation splits. An extensive set of experiments was conducted, followed by a comprehensive statistical analysis.

The results demonstrate the superiority of the simple IDW method over the other reference methods in terms of both reconstruction accuracy and computational efficiency. IDW achieved the lowest RMSE ($3.00 \pm 1.93$), MAE ($1.32 \pm 0.77$), and $\Delta_{MAX}$ ($24.06 \pm 17.15$), as well as the highest $R^2$ ($0.68 \pm 0.16$), across 100 randomly sampled sparse datasets from the ECA\&D database. Differences in RMSE, MAE, and $R^2$ were statistically significant and exhibited moderate to large effect sizes. The Dunn post-hoc test further confirmed the consistent superiority of IDW across all evaluated quality measures.

The evaluation of reconstruction time and peak memory usage for different reconstruction domain sizes showed that IDW required less computational time (even when excluding the training process of MMGN). However, memory usage for both IDW and OK increased more rapidly with the number of target points, which aligns with expectations since IDW and OK store values explicitly, whereas MMGN provides an implicit representation.

The experiments highlight that methods with a small number of interpretable hyper-parameters, such as IDW, are considerably easier to optimize and more robust than complex approaches like the MMGN artificial neural model. We conclude that the use of artificial neural networks should always be critically assessed for each specific application, balancing their potential advantages against the substantial computational cost and complexity of the hyper-parameter optimization process.
\end{abstract}

\begin{IEEEkeywords}
spatial reconstruction, climate data approximation, inverse distance weighting, ordinary kriging, implicit neural representation,
\end{IEEEkeywords}

\section{Introduction}

\IEEEPARstart{O}{bservations} play a pivotal role in advancing climate modelling, particularly in the context of climate change research \cite{guo2015earth}, extreme event attribution \cite{christidis2013new}, and operational oceanography \cite{davidson2019synergies}. Accurate observational data enable model calibration, facilitate bias correction \cite{eyre2016observation}, and ensure reliable quality assessment of climate models \cite{gomez2012role,tapiador2017global,eyring2019taking}, forming the backbone of climate science research.

Observational data come mainly from two main sources: remotely sensed data and in-situ measurements. Remotely sensed observations, such as those obtained via satellites \cite{crichton2012sharing}, aircraft \cite{zhang2020estimation}, drones \cite{barfuss2023drone}, radiosondes \cite{ingleby2016progress}, or ground-based devices such as Doppler radar systems \cite{lindskog2004doppler}, provide extensive spatial coverage and high temporal resolution. On the other hand, in-situ observations, representing localized measurements of physical fields, offer higher accuracy and are invaluable for validation. To overcome the limitations of individual datasets, both types are often integrated \cite{arabi2020integration}.

Despite their value, in-situ observations face significant challenges. These include measurement biases, uneven spatial distribution, and persistent gaps, especially in regions with limited infrastructure or harsh conditions \cite{bessenbacher2023optimizing}. These observational sparsity hinder their direct application in climate models as to achieve spatial and temporal continuity the data needs to be spatially reconstructed \cite{bessenbacher2023gap}.

In this study, we address the challenge of sparsity in in-situ observations, focusing specifically on temperature as a critical variable in climate modelling due to its central role in atmospheric processes. We carefully design experiments to reliably assess the performance of three benchmark methods of varying complexity under optimised hyper-parameters. Our findings clearly demonstrate that simple methods can, in some cases, be more accurate and easier to apply in practice, which challenges the current trend of carelessly applying artificial neural networks to a wide range of problems.

Another contribution of this work is the integration of hyper-parameter optimisation, based on \textit{optuna} \cite{akiba2019optuna}, into the \textit{Climatrix} tool \cite{climatrix}.

\section{Related Studies}
Artificial intelligence is already widely applied across various domains of climate sciences \cite{dewitte2021artificial}. In particular, AI has been extensively exploited in processes such as data assimilation \cite{boukabara2020artificial}, bias reduction \cite{schneider2017earth}, land cover classification \cite{helber2019eurosat}, and retrieving geophysical information from remotely sensed observations \cite{krasnopolsky2003some}. Recent advancements have also achieved impressive results in accurate weather forecasting \cite{lam2023learning} and extreme event prediction \cite{nearing2024global}.

Although data-driven methods have been employed for climate data reconstruction for many years, new approaches relying on implicit neural representations (INRs) are now emerging with hope to address limitations of traditional techniques.

We define the problem of \textit{reconstruction} in this study as the task of identifying a function $\mathcal{F}: \mathbb{R}^2 \rightarrow \mathbb{R}$ that approximates a given set $\mathcal{X}$ of 3D points $\mathbf{x}_i \in \mathcal{X}$, where $\mathbf{x}_i = \{x_i^{(1)}, x_i^{(2)}, x_i^{(3)}\}$, and satisfies the following condition:

\begin{equation}
\label{eq:recons}
\forall_{\mathbf{x} \in \mathcal{X}} \mathcal{F}(\mathbf{x}^{(1)}, \mathbf{x}^{(2)}) = \mathbf{x}^{(3)} + \epsilon(\mathbf{x}),
\end{equation}

where $x^{(j)}$ is the $j$-th coordinate of a point $\mathbf{x} \in \mathcal{X}$, and $\epsilon(\mathbf{x})$ represents an approximation error that is to be minimized. For interpolation, this error is strictly zero: $\epsilon(\mathbf{x}) = 0$ for reconstruction nodes.

\subsection{Traditional Reconstruction Methods}
\subsubsection{Spatial Interpolation}
The interpolation task involves finding a function that exactly matches the prescribed values of a dataset \cite{franke1982scattered}, i.e., $\epsilon(\mathbf{x}) = 0$ for reconstruction nodes. Depending on the data's characteristics, the underlying function may be subject to constraints or prior assumptions.

A variety types of interpolation methods exist, two of which are particularly noteworthy. The first relies on Lagrange polynomials \cite{Scherer2013}, enabling the construction of an $(b-1)$-degree polynomial (where $b = |\mathcal{X}|$) that passes through $b$ points. This method can also be extended to multivariate data \cite{sauer1995multivariate}.

The second and more widely adopted approach in climate sciences is built on the intuition that closer reconstruction nodes have a greater influence than distant ones. This category includes methods such as Inverse Distance Weighting (IDW) and kriging \cite{ozelkan2015spatial}. For an arbitrary point $\bar{\mathbf{x}}$ located within the bounds of reconstruction nodes, the reconstruction function in IDW takes the following form:

\begin{equation}
\label{eq:idw}
\mathcal{F}(\bar{\mathbf{x}}^{(1)}, \bar{\mathbf{x}}^{(2)}) = \frac{\sum_{i < |\mathcal{X}|} \mathbf{x}_i^{(3)} w_i}{\sum w_i},
\end{equation}

where $w_i$ is the reciprocal of the powered (usually squared) distance between a reconstructed point and a reconstruction node.

For kriging methods, which often outperform IDW \cite{zimmerman1999experimental}, node weights are determined differently. They incorporate the spatial autocorrelation of the data modelled using a semivariogram. Traditional kriging assumes the data is both mean stationary and isotropic. However, advanced kriging variants allow some of these assumptions to be relaxed.

\subsubsection{Sub-grid Parametrization}
Many climate phenomena occur at spatial and temporal scales smaller than the resolution of physical models, making them challenging to capture directly. Sub-grid parametrisation addresses this gap \cite{fielding2020parametrizing,yuval2020stable}. This approach is widely utilized in climate science to enhance the fidelity and applicability of physical models.

Sub-grid parametrisation involves tuning (and calibrating) physics-based models to better align with specific scientific objectives or observational datasets \cite{hourdin2017art}. The authors of \cite{hourdin2017art} identify two main aspects of this process:

\begin{enumerate}
    \item \textbf{Objective tuning} --- This involves optimizing parameters through well-defined mathematical approaches, such as solving Bayesian calibration problems \cite{kennedy2001bayesian}. 
    \item \textbf{Subjective judgement} --- This aspect refers to expert decisions made during the parametrization process, such as prioritizing specific processes or choosing parameter values based on domain knowledge, expertise, and scientific priorities. 
\end{enumerate}

By fusing these approaches, sub-grid parametrization enables climate models to approximate the set of observations, provided the physical model is supplied.

\subsection{Deep Learning Methods} 
Deep learning models have emerged as powerful tools for data reconstruction problems in climate science \cite{wang2023reconstruction}, owing to their ability to capture hidden patterns in data through data-driven approach.

The first category of deep-learning-based methods are super-resolution techniques. They involve upscaling or enhancing the resolution of gridded datasets \cite{nguyen2001computationally}, which are typically represented as images or similarly structured data formats. In recent years, machine learning, particularly deep learning, has shown impressive capabilities in super-resolution tasks. These methods have become increasingly popular across diverse fields, with artificial neural networks being deployed to significantly improve resolution enhancement \cite{ma2019image,wang2020hyperspectral}. These group of solutions, however, operate on gridded  (image-like) structures, hence their usability for sparse in-situ observations is limited and doubtful.

For super-resolution, convolutional neural networks (CNNs) are frequently employed due to their ability to effectively capture spatial patterns and features in image-like data. In addition to CNNs, generative models such as generative adversarial networks (GANs) are also utilized to further improve super-resolution outcomes \cite{creswell2018generative, jiang2019edge}.

A particularly promising and versatile approach in the domain of deep learning is the use of \textbf{implicit neural representations} (INRs). Despite potential confusion surrounding the term \textit{implicit}, INR methods can be expressed both in an implicit form (\ref{eq:inr}) and an explicit form (\ref{eq:inr2}):

\begin{equation}
    \label{eq:inr}
    \mathcal{F}_\theta(\mathbf{x}^{(1)}, \mathbf{x}^{(2)}, \mathbf{x}^{(3)}) \approx 0,
\end{equation}

\begin{equation}
    \label{eq:inr2}
    \mathcal{F}_\theta(\mathbf{x}^{(1)}, \mathbf{x}^{(2)}) \approx \mathbf{x}^{(3)},
\end{equation}

where \(\mathcal{F}_\theta\) denotes the implicit neural representation, and \(\mathbf{x}^{(i)}\) are components of a sample in a multidimensional space.

In this context, the term \textit{implicit} does not refer to the mathematical form of the function itself, but rather to the way the signal is stored. Unlike methods such as IDW or kriging, which rely on discrete memory structures, INRs encode information implicitly through the parameterisation defined by the neural network’s weights \cite{sitzmann2020implicit}.

INRs have been successfully applied in various fields, including 3D shape reconstruction \cite{ye2022gifs}, where the signed distance function serves as the reconstruction objective.

One of the most recent methods proposed for sparse data reconstruction is the MMGN model introduced by \cite{luo2024continuous}. This approach employs an encoder–decoder architecture enhanced with iterative Gabor-based feature extraction. The authors report superior performance, in terms of mean squared error, compared to several established reconstruction methods, including the widely recognised SIREN architecture \cite{sitzmann2020implicit}. For this reason, we selected MMGN as the reference deep learning method in our study.

\section{Methods}
\subsection{Reference methods}
As reference methods, we selected three well-known and widely applied reconstruction approaches of varying complexity: 

\begin{enumerate}
    \item IDW --- a simple method with only three hyper-parameters, easy to tune, belonging to the class of explicit representations (all reconstruction nodes are stored),
    \item ordinary kriging --- the most common variant of kriging \cite{wackernagel2003ordinary}, conceptually similar to IDW but with weights computed in a more sophisticated manner,
    \item MMGN --- a recent INR-based method reported to achieve superior performance in data reconstruction compared to others in its family.
\end{enumerate}

\subsection{Definitions}
Sparse climate observations can be thought of as a point cloud, yet with characteristics that differ slightly from the conventional understanding of 3D point clouds typically utilized in volumetric-data studies \cite{fei2022comprehensive}. To account for these differences, the definition of a point cloud can be extended from a simple collection of Cartesian coordinates to a representation in any metric space \cite{bello2020deep}, which is not necessarily constrained to three dimensions. Hence, we defined the climate-based point cloud as expressed in (\ref{eq:pc})-(\ref{eq:pt}).

\begin{equation}
    \label{eq:pc}
    \mathcal{P} = \{p_i~~|~~i \in \mathbf{I}\},
\end{equation}
where $\mathbf{I}$ is a set of indices, $\mathbf{I} = \{1, 2, 3, ..., N\}$, representing a collection of points whose size is given by $|\mathbf{I}| = |\mathcal{P}| = N$.

Each point in a climate-based point cloud can therefore be defined as a tuple that combines its geographic position with the corresponding physical field values. For our study, we specifically considered a single physical field: \textbf{temperature}. This formulation is expressed in (\ref{eq:pt}). 

\begin{equation}
    \label{eq:pt}
    p_i = (\lambda, \phi, \tau),
\end{equation}

where $\phi$ represents the longitude, $\lambda$ denotes the latitude, and $\tau$ corresponds to the physical field, in this case, the temperature at the geographic coordinates $(\lambda, \phi)$. 

\begin{figure}
    \centering
    \includegraphics[width=0.55\textwidth]{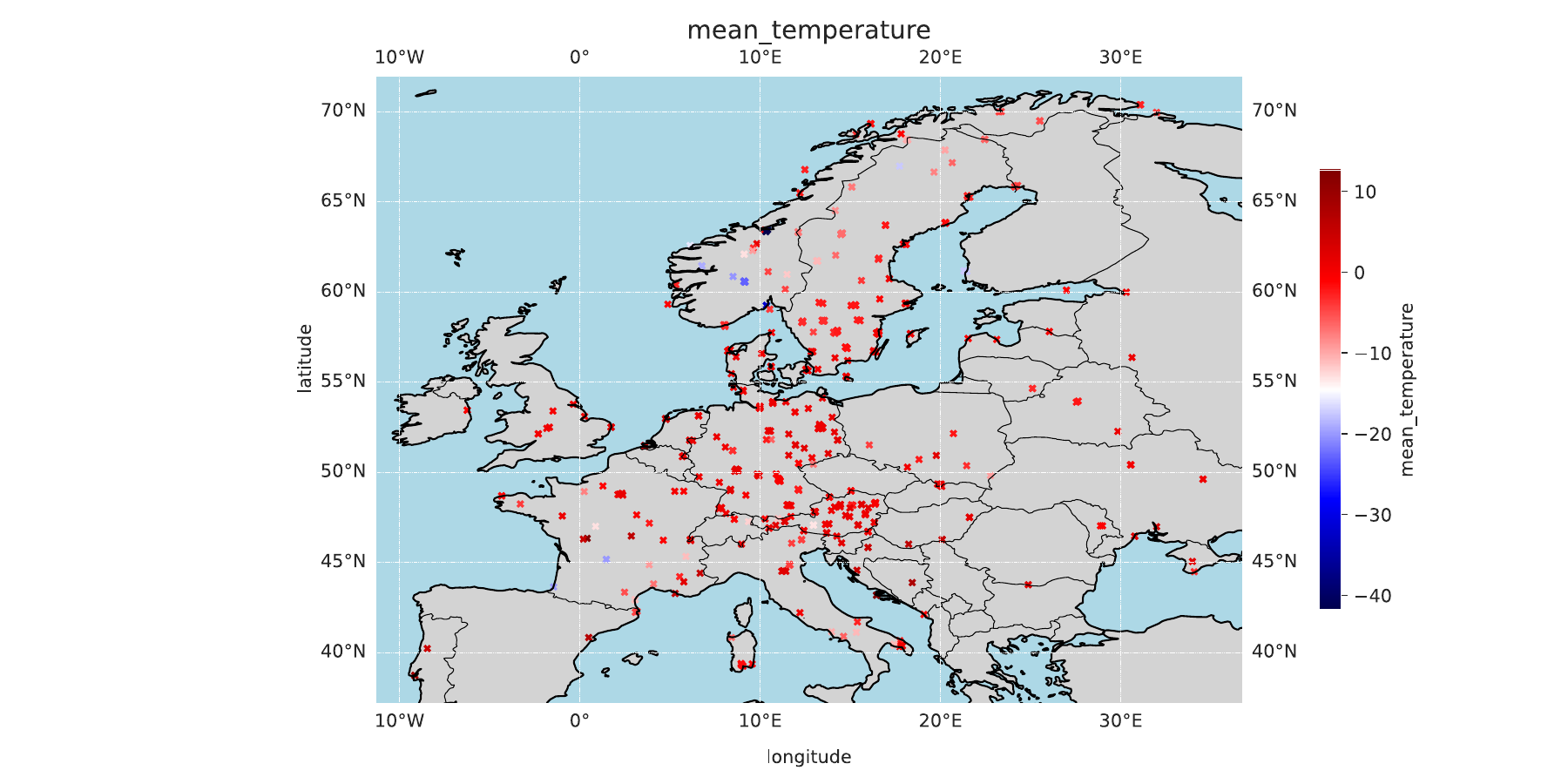}
    \caption{A sample climate point cloud representing temperature field}
    \label{fig:cpc}
\end{figure}

Let us confront that concept with the general notation introduced in the section Related Studies --- the climate-based point cloud points $p_i \in \mathcal{P}$ are treated as the reconstruction nodes $\mathbf{x}_i \in \mathcal{X}$ ($\mathcal{P} = \mathcal{X})$. A sample climate point cloud is represented in Fig. \ref{fig:cpc}.

\subsection{Tools used}
In our study, we used the \textit{Climatrix} \cite{climatrix} accessor to facilitate and accelerate climate data manipulation and to enable systematic comparison of reconstruction methods.

\subsection{Quality Metrics}
In our study, we adopted a robust, evidence-based methodological framework for evaluation, as proposed in \cite{walczak2025evidence}.

To evaluate the quality of the reference climate reconstruction methods, we employed three complementary quality measures: root mean square error ($RMSE$), mean absolute error ($MAE$), and the determination coefficient ($R^2$). These metrics offer distinct insights into the method’s performance. 

\begin{enumerate}
    \item \textbf{Root mean squared error} (\ref{eq:rmse}) --- this metric measures error in the same units as the observations, providing an interpretable indication of overall predictive accuracy.

    \begin{equation}
    \label{eq:rmse}
    RMSE = \sqrt{\frac{1}{N} \sum_{i=1}^{N} (\tau_i - \hat{\tau}_i)^2},
\end{equation}
where $N$ represents the cardinality of the climate point cloud, $\tau_i$ denotes the observed temperature at point $p_i$, and $\hat{\tau}_i$ is the predicted temperature derived from the approximation function.

\item \textbf{Mean absolute error} (\ref{eq:mae}) --- this metric calculates the average absolute difference between observed and predicted values, making it more robust against outliers than $RMSE$.

\begin{equation}
    \label{eq:mae}
    MAE = \frac{1}{N} \sum_{i=1}^{N} |\tau_i - \hat{\tau}_i|,
\end{equation}

\item \textbf{Determination coefficient} (\ref{eq:r2}) according to \cite{chicco2021coefficient} is more informative than RMSE and MAE due to its upper bound set to $1$.

\begin{equation}
    \label{eq:r2}
    R^2 = 1 - \frac{\sum (\tau_i - \hat{\tau}_i)^2}{\sum (\tau_i - \bar{\tau}_i)^2},
\end{equation}
where $\bar{\tau}$ is the mean value of the observed temperature.
\end{enumerate}

Additionally, we report the maximum absolute difference $\Delta_{MAX}$ (\ref{eq:mdiff}) between observed and predicted temperature values. This metric highlights the worst-case prediction error, providing insight into each method’s behaviour under extreme conditions.

\begin{equation}
    \label{eq:mdiff}
    \Delta_{MAX} = \max_{i \in \mathbf{I}} |\tau_i - \hat{\tau}_i|
\end{equation}

To evaluate the efficiency of the reconstruction methods, we further measured for simulated target data both the elapsed time $t_r(M)$ and  the maximum memory occupancy $m_r(M)$ required to reconstruct $M \in \{10^2, 5 \cdot 10^2, 5 \cdot 10^2, 10^3, 5 \cdot 10^3, 10^4, 5 \cdot 10^4, 10^5, 5 \cdot 10^5, 10^6 \}$ points.

\subsection{Dataset}
At the beginning, it is important to emphasize a fundamental limitation: \textbf{a ground-truth dataset of dense, in-situ observations does not exist}. Therefore, to evaluate the effectiveness of the proposed method, we rely on blended in-situ observations provided by the European Climate Assessment \& Dataset (ECA\&D) project \cite{klein2002daily}, which serves as the basis for the gridded E-OBS dataset \cite{cornes2018ensemble}.

The ECA\&D project, a continuation of the original ECA initiative coordinated by the Royal Netherlands Meteorological Institute (KNMI), focuses on analysis of \textbf{daily} precipitation and temperature (particularly trends in extreme values \cite{eca2013european}) records from meteorological stations across Region VI (Europe), as defined by the World Meteorological Organization (WMO) \cite{wmo_regions}. The dataset is provided in two formats: non-blended, which contains raw observations from partner institutions, and blended, where missing values are filled using observations from nearby meteorological stations (within 12.5 km and with a relative elevation difference no greater than 25 m) \cite{eca2013european}.

Due to its collaborative nature, the ECA\&D dataset exhibits spatial and temporal heterogeneity in coverage.

In this study, we use the blended version, which underpins the widely adopted E-OBS gridded climate product.

To ensure seasonal and spatial representativeness, and to enable robust statistical analysis, we randomly selected $100$ ($|\mathcal{S}| = 100$) single dates across the entire ECA\&D archive to form a dataset $\mathcal{S}$. We considered only those dates where number of valid observations with the quality flag set to \texttt{valid} exceeds $500$. The selection was performed using the \texttt{NumPy} \cite{harris2020array} pseudo-random number generator with a fixed seed of $0$ to ensure reproducibility.

We chose this relatively large number of dates to capture variability in spatial station density and to mitigate potential biases introduced by anomalous weather events or extreme conditions.

For each selected date, all available in-situ observations were randomly divided into three mutually exclusive subsets: training, validation, and testing. 

By focusing on single-day snapshots, we eliminate temporal dependencies, thereby reducing the risk of data leakage --- particularly the 'neighbourhood bias` described in \cite{kapoor2024reforms}. Each single-date sample serves as an independent input to the instantaneous reconstruction task.

Across all dates, 60\% of the available observations were assigned to the training set, with the remaining 40\% equally and randomly split between the validation and test sets, all derived from the same day's data. Cardinalities of splits may vary across dates due to time-dependent differences in spatial coverage in ECA\&D dataset. Summary statistics for the training and validation splits are shown in Table~\ref{tab:stats}. To prevent data leakage and unintentional methodological bias, we did not compute statistics for the test split.

\begin{table}[]
\caption{Summary statistics for training and validation splits across 100 randomly selected dates.}
\label{tab:stats}
\centering
\begin{tabular}{|c|c|c|c|c|}
\hline
\textbf{Split}      & \textbf{Min [$^oC$]} & \textbf{Mean [$^oC$]} & \textbf{St. dev. [$^oC$]} & \textbf{Max [$^oC$]} \\ \hline
\textbf{Train}      &      $-57.00$        &  $6.65$             &  $10.41$                 &               $37.60$\\ \hline
\textbf{Validation}  &      $-53.40$        &  $6.68$             &  $10.44$                 &               $37.60$\\ \hline
\end{tabular}
\end{table}

\subsection{Experiment Procedure}
\subsubsection{Reproducibility}
To ensure full reproducibility of the experiments described in this paper, we manually set seeds for pseudo-random number generators when processing each individual date. 

Information regarding library versions and specific implementation details is available in the \texttt{experiments} directory of the \textit{Climatrix} repository we contributed to\footnote{\textit{Climatrix} project repository: \url{https://github.com/jamesWalczak/climatrix}.}.

\subsubsection{Dataset preprocessing}
Based on exploratory data analysis, we found no need for additional data cleaning or imputation.

However, we applied method-specific preprocessing. For ordinary kriging, we normalised latitude and longitude to the range $[-1, 1]$ and standardised the physical field values by subtracting the mean and dividing by the standard deviation.  

IDW does not require any preprocessing.  

For the MMGN method, we applied min–max scaling to the range $[-1, 1]$, following the original setup \cite{luo2024continuous} so that the numerical problem with float overflowing can be avoided.

\subsubsection{Experimental trial configuration and split setup}
Our investigation focused on analysing the quality and efficiency of the three optimised reference reconstruction methods.  

We sampled $100$ time instants from the ECA\&D dataset and trained the selected methods using predefined training splits. Validation splits were used to compute the optimisation criterion for hyper-parameter tuning (see Section \ref{sec:hpopt}). Test splits were held out and used exclusively to report performance for the tuned methods at each sampled time instant.

After applying the optimised hyper-parameters, we reconstructed values for all spatial points in the test split. Each reconstruction method produced a set of reconstructed values $\hat{\mathcal{T}}$ over the same spatial domain as the test split. The elements of this set are tuples $r_i = (\lambda_i, \phi_i, \hat{\tau}_i)$, where $\lambda_i, \phi_i$ are the geographical coordinates of the $i$-th test point.

To investigate reconstruction times for the three reference methods, we used the optimised hyper-parameters obtained for each sample throughout the experiment. The training split of each sample served as the set of reconstruction nodes, while synthetically generated datasets were used as targets to be reconstructed. These target datasets domains were sampled from a uniform distribution (taking as limits the minimum and maximum values of latitude and longitude or a training dataset) with increasing cardinalities of $\{10, 100, 500, 1000, 2000, 5000, 10000\}$ points. For MMGN, we neglected time necessary for training a model (we assumed the weights are preloaded in advance).

\subsection{Hyper-parameter optimisation}
\label{sec:hpopt}
To facilitate hyper-parameter optimisation, we extended the \textit{Climatrix} tool with the \texttt{HParamFinder} component, which relies on the \textit{optuna} library.

For optimisation, we adopted Gaussian process priors for Bayesian optimisation \cite{snoek2012practical}. Although Bayesian optimisation does not guarantee convergence to the global optimum due to its probabilistic nature, brute-force optimisation is often impractical for multidimensional space of numerical hyper-parameters.

Table \ref{tab:bo_setup} presents the optimisation setup for all three reference methods.

\begin{table}[H]
\caption{Bayesian optimisation setup for reference methods}
\label{tab:bo_setup}
\centering
\begin{tabular}{|p{2cm}|p{2cm}|p{2cm}|}
\hline
\textbf{Method} &
\textbf{\# initial samples} & \textbf{\# iterations} \\ \hline
IDW & 50 & 100 \\ \hline
OK & 50 & 100 \\ \hline
MMGN & 50 & 200 \\ \hline
\end{tabular}
\end{table}

Because MMGN has many more hyper-parameters to optimise, we allocated twice as many trials as for IDW or OK.

For each method, we empirically set constraints on the hyper-parameters to ensure that most optimised values fall within, but not at the boundaries of, the imposed ranges. To verify this, we tracked histograms of continuous hyper-parameters and frequencies of categorical choices.

As the optimisation criterion, we minimised the Mean Absolute Error (MAE) (\ref{eq:mae}) on the validation split of each sample.

\subsubsection{Bounds for IDW}
IDW requires only two hyper-parameters: the number of nearest neighbours and the power coefficient, which controls how rapidly influence decreases with distance. Their bounds are given in Table \ref{tab:bo_idw}.

\begin{table}[H]
\caption{Optimisation bounds for IDW}
\label{tab:bo_idw}
\centering
\begin{tabular}{|p{2cm}|p{2cm}|p{2cm}|}
\hline
\textbf{Parameter} & \textbf{Type} & \textbf{Values/Range} \\ \hline
 \# nearest neighbours & integer & $[1, 50]$  \\ \hline
 Distance power & real number & $[10^{-7}, 5.0]$  \\ \hline
\end{tabular}
\end{table}

\subsubsection{Bounds for OK}
Ordinary kriging involves several hyper-parameters with bounds listed in Table \ref{tab:bo_ok}.

\begin{table}[H]
\caption{Optimisation bounds for ordinary kriging}
\label{tab:bo_ok}
\centering
\begin{tabular}{|p{2cm}|p{2cm}|p{3cm}|}
\hline
\textbf{Parameter} &  \textbf{Type} &\textbf{Values/Range} \\ \hline
 \# averaging bins for semivariogram & integer & $[2, 50]$  \\ \hline
 Anisotropy scaling factor & real number & $[10^{-5}, 5]$ \\ \hline
 Coordinate type & categorical & \{euclidean, geographic\}  \\ \hline
 Semivariogram model & categorical & \{linear, power, gaussian, spherical, exponential, hole-effect\} \\ \hline
\end{tabular}
\end{table}

\subsubsection{Bounds for MMGN}
MMGN is a neural network model, and therefore involves both generic and architecture-specific hyper-parameters. Their bounds are shown in Table \ref{tab:bo_mmgn}. We fixed the number of epochs at $500$, during which the optimisation criterion was minimised.

\begin{table}[H]
\caption{Optimisation bounds for MMGN}
\label{tab:bo_mmgn}
\centering
\begin{tabular}{|p{2cm}|p{2cm}|p{3cm}|}
\hline
\textbf{Parameter} &  \textbf{Type} &\textbf{Values/Range} \\ \hline
 learning rate & real number & $[10^{-5}, 10^{-2}]$  \\ \hline
 L2 regularisation & real number & $[0, 0.1]$ \\ \hline
 mini-batch size & integer & $[32, 1024]$ \\ \hline
 \# hidden dimensions & categorical & \{32, 64, 128, 256, 512, 1024\} \\ \hline
 \# latent dimensions & categorical & \{32, 64, 128, 256, 512, 1024\}  \\ \hline
 \# layers & integer & $[1, 10]$ \\ \hline
 input scale & real number & $[2, 1024]$ \\ \hline
 alpha & real number & $[0.0, 100.0]$ \\ \hline
\end{tabular}
\end{table}

Here, the learning rate controls the step size during gradient descent, while L2 regularisation mitigates overfitting. Hidden and latent dimensions govern the internal feature representations. The input scale is a parameter of the Gabor filter, and $\alpha$ is the concentration parameter of the gamma distribution used in the filter. For details, see the code accompanying MMGN \cite{luo2024continuous}.

\subsubsection{Statistical considerations}
To evaluate the outcomes of our experiments, we performed a series of statistical analyses. In particular, we applied the non-parametric Kruskal–Wallis test \cite{kruskal1952nonparametric,kruskal1952use} to assess the impact of reconstruction method on each reported quality measure. When the Kruskal–Wallis test indicated significance, we performed post hoc analyses using Dunn’s test \cite{dunn1964multiple} with Holm–Bonferroni correction for multiple comparisons \cite{holm1979simple}. 

We also reported effect sizes. For the Kruskal–Wallis test, we computed $\eta^2$ following \cite{tomczak2014need}:

\begin{equation}
    \label{eq:es_kw}
    \eta^2= \frac{H - k + 1}{k\cdot|\mathcal{S}|-k},
\end{equation}
where $H$ is the Kruskal–Wallis statistic, $|\mathcal{S}|=100$, and $k$ is the number of groups --- in our case set to three (as we have three methods: IDW, OK, and MMGN).

For pairwise Dunn’s tests, we calculated the biserial rank correlation coefficient $r$ \cite{tomczak2014need}:

\begin{equation}
    \label{eq:es_dt}
    r = \frac{2 \cdot (\bar{R_1} - \bar{R_2})}{n_1+n_2},
\end{equation}
where $\bar{R_1}$ and $\bar{R_2}$ are the mean ranks of groups 1 and 2, and $n_1$ and $n_2$ are their respective sizes. In our case, $n_1=n_2=|\mathcal{S}|=100$, as we sampled $100$ instances from the ECA\&D dataset.

For both Kruskal–Wallis and Dunn’s tests, we set the significance threshold at $\alpha=0.05$.

\section{Results}
Table~\ref{tab:res_overall} presents the aggregated quality measures for the three examined reference reconstruction methods across all test splits of $\mathcal{S}=100$ samples. Reported values correspond to the median followed by the interquartile range (as distributions are strongly abnormal). 

\begin{table}[]
\caption{Quality measure values for the three evaluated methods across all test splits of $|\mathcal{S}|=100$ samples. Each entry shows the median followed by the interquartile range. Green cells indicate the best value of a given metric, while red cells indicate the worst.}
\label{tab:res_overall}
\centering
\begin{tabular}{|c|c|c|c|}
\hline
\textbf{Metric} & \textbf{IDW} & \textbf{OK} & \textbf{MMGN} \\
\hline
RMSE   &   \cellcolor{green!25} $ 3.00 \pm 1.93 $ & \cellcolor{red!25} $ 5.40 \pm 3.30$  &  $4.11 \pm 2.46$   \\ \hline
MAE    &  \cellcolor{green!25} $ 1.31 \pm 0.77 $  & \cellcolor{red!25}  $3.81 \pm 2.33$  &   $1.93 \pm 1.02$  \\ \hline
$R^2$  &  \cellcolor{green!25} $0.67 \pm 0.16$ &  \cellcolor{red!25} $-0.76 \pm 1.92$  &  $0.46 \pm 0.21$  \\ \hline 
$\Delta_{MAX}$     & \cellcolor{green!25} $24.06 \pm 17.15$ &  $24.85 \pm 17.40$   & \cellcolor{red!25} $35.83 \pm 21.46$   \\ \hline
\end{tabular}
\end{table}

Following the established methodology, we performed an omnibus Kruskal–Wallis test for each metric separately. The results are reported in Table~\ref{tab:kruskal_wallis}.

\begin{table}[]
\caption{Results of the omnibus Kruskal–Wallis test for all quality measures.}
\label{tab:kruskal_wallis}
\centering
\begin{tabular}{|c|c|c|c|}
\hline
\textbf{Metric} & $H$ \textbf{statistic} & $p$-\textbf{value} & \textbf{Effect size}\\
\hline
RMSE   &   $70.67$ &  $4.50 \cdot 10^{-16}$ & $0.23$ \\ \hline
MAE    &  $169.66$ & $1.44 \cdot 10^{-37}$  & $0.56$\\ \hline
$R^2$  &   $288.95$ & $4.3\cdot 10^{-46}$ & $0.70$\\ \hline 
$\Delta_{MAX}$     & $28.24$ & $7.39 \cdot 10^{-7}$ & $0.01$  \\ \hline
\end{tabular}
\end{table}

For all quality measures, the $p$-values are far below the presumed significance threshold ($0.05$), indicating strong evidence of statistically significant differences between the methods. The effect sizes are substantial for all metrics except $\Delta_{MAX}$, where the value is negligible.

The complete results of the Dunn post-hoc tests are provided in the associated experiment notebook\footnote{The notebook with a detailed analysis is available at \url{https://github.com/jamesWalczak/climatrix/blob/main/experiments/jwalczak/01_Apr_02_compare_recon_method/notebooks/analysis.ipynb}}. In brief, $p$-values for RMSE, MAE, and $R^2$ are well below $0.05$, confirming statistical significance. For $\Delta_{MAX}$, significant differences were also observed, except for the IDW vs. OK comparison, where $p=0.32$. Considering the median values of all reported quality measures across the three methods (IDW, OK, MMGN), the results clearly demonstrate the superiority of IDW over both OK and MMGN in terms of RMSE, MAE, and $R^2$ when reconstruction proceeds the hyper-parameters optimisation. 

In Table~\ref{tab:dunns_effect_size}, the $r$ biserial rank correlation coefficients are presented for the Dunn post-hoc tests.

\begin{table}[htbp]
\centering
\caption{Effect size for Dunn post-hoc tests for all reported measures}
\label{tab:dunns_effect_size}
\begin{tabular}{|c|c|c|}
\hline
\textbf{Metric} & \textbf{Pair} & $r$ \\
\hline
\multirow{3}{*}{RMSE} & IDW vs OK& $-0.65$ \\ \cline{2-3}
                       & IDW vs MMGN& $-0.41$\\ \cline{2-3}
                       & OK vs MMGN & $0.35$\\ 
\hline
\multirow{3}{*}{MAE}& IDW vs OK& $-0.95$ \\ \cline{2-3}
                       & IDW vs MMGN& $-0.48$\\ \cline{2-3}
                       & OK vs MMGN & $0.78$\\
\hline
\multirow{3}{*}{$R^2$} & IDW vs OK& $0.99$ \\ \cline{2-3}
                       & IDW vs MMGN& $0.64$\\ \cline{2-3}
                       & OK vs MMGN & $-0.91$\\
\hline
\multirow{3}{*}{$\Delta_{\max}$} & IDW vs OK& $-0.08$\\ \cline{2-3}
                       & IDW vs MMGN& $-0.41$\\ \cline{2-3}
                       & OK vs MMGN & $-0.33$\\
\hline
\end{tabular}
\end{table}

As shown in Table~\ref{tab:dunns_effect_size}, the effect sizes for RMSE, MAE, and $R^2$ are large across all pairwise comparisons, favouring IDW over both MMGN and OK (negative values for error measures and positive for $R^2$). For $\Delta_{max}$, the reported effect sizes are moderate (for IDW vs. MMGN and OK vs. MMGN) or small (for IDW vs. OK).

\begin{figure}
\centering
\includegraphics[width=0.9\linewidth]{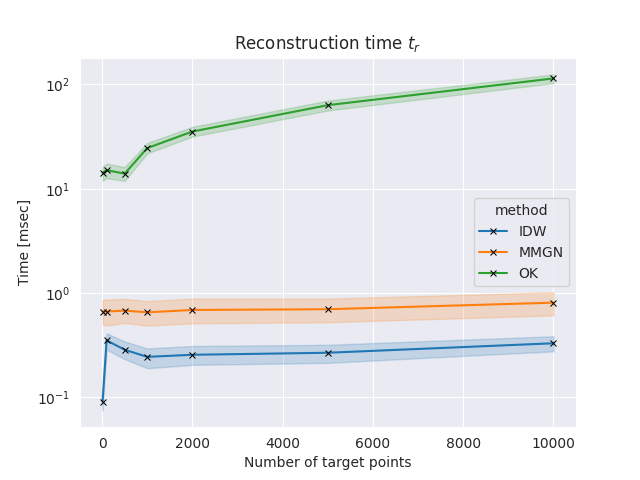}
\caption{Reconstruction time for optimized hyper-parameters with an increasing number of points to be reconstructed.}
\label{fig:time_r}
\end{figure}

\begin{figure}
\centering
\includegraphics[width=0.9\linewidth]{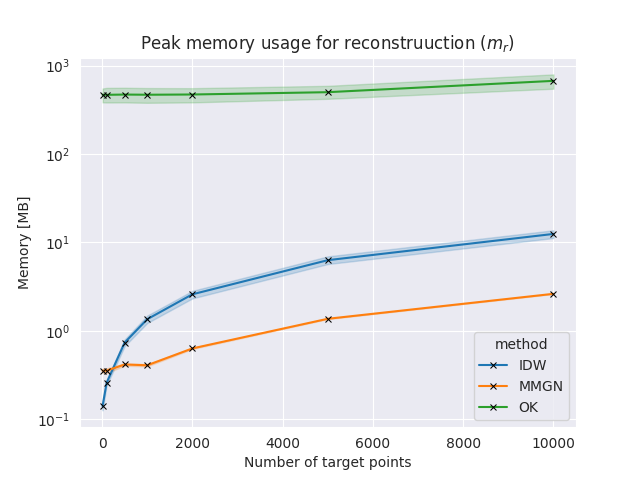}
\caption{Peak memory usage during reconstruction for optimized hyper-parameters with an increasing number of points to be reconstructed.}
\label{fig:mem_r}
\end{figure}

We now proceed to compare the computational efficiency of the methods in terms of reconstruction time, $t_r$ (Fig.~\ref{fig:time_r}), and peak memory usage, $m_r$ (Fig.~\ref{fig:mem_r}). Due to the large differences in the required time and memory consumption for ordinary kriging, a logarithmic scale was applied to the y-axis. For all three reference methods, we depicted the $95^{th}$ confidence intervals for both time and memory.

Although the IDW method belongs to the class of explicit representations and therefore retains all training points in memory, there is no noticeable difference in the order of reconstruction time for up to $10,000$ points. Interestingly, the IDW method requires less time than MMGN across all examined target domain sizes and significantly less than OK, whose time and memory demands are far beyond comparison. The short reconstruction time of the IDW method is particularly evident for sparse yet moderately sized training sets, where efficient access is provided by the kd-tree structure used for nearest-neighbour retrieval. For noticeably larger training sets, however, the difference between IDW and MMGN is expected to shift in favour of the implicit neural representation.

For MMGN, which is an implicit neural representation method, the poorer-than-expected performance results from the overhead associated with artificial neural network processing. The most critical factors include memory transfers between CPU and GPU, data batching, and data denormalisation required to match the expected input range. It should also be emphasized that for MMGN the reported time corresponds solely to the reconstruction phase; the model training time is not included and may further impact the overall computational efficiency of the pipeline.

\section{Discussion and Conclusions}
In this study, we addressed the problem of reconstructing continuous physical fields from sparse \textit{in-situ} observations using three reference methods: inverse distance weighting (IDW), ordinary kriging (OK), and an implicit neural representation model, MMGN. Based on careful and exhaustive experiments, we conclude that, after hyper-parameter optimization, the simplest method (IDW) yields the best results among all three reference methods across all evaluated quality measures. This superiority is statistically significant and exhibits a large effect size for RMSE, MAE, and $R^2$.

We also found that IDW, though reliant on distance computations and an explicit in-memory representation of the training set, requires considerably less time to perform reconstruction. This advantage comes at the cost of a somewhat higher, though roughly constant-factor, peak memory consumption.

Our findings challenge the widespread belief in the universal applicability of artificial neural networks for all types of problems. We wish to emphasize that we do not deny the high capabilities of ANNs; however, they are tools whose selection should be guided by the nature of the problem at hand. In some cases, their complexity and flexibility may be unnecessary or even problematic. We do not claim that ANNs (and in this particular case, the MMGN method) are incapable of efficiently solving the stated problem. In fact, ANNs possess remarkable potential, which is often difficult to fully exploit due to the high dimensionality of their hyper-parameter space and the complexity of optimization.

While IDW relies on only two easily interpretable parameters, MMGN involves a significantly larger number of hyper-parameters, some of which could not be thoroughly explored due to computational constraints and time limitations. The key conclusion from our experiments is that simpler methods can often outperform more advanced ones, not necessarily because of their limited capabilities, but because the optimization of complex models is constrained by the vastness of their hyper-parameter space.

Furthermore, it is worth noting the computational efficiency of the compared methods. In our case, where the task involves reconstructing sparse spatial points rather than processing massive datasets, the in-memory (explicit) representation inherent to IDW requires considerably fewer computational resources than the implicit neural representation used in MMGN.

Finally, it should be noted that hyper-parameters were optimized using validation splits, which may not always ensure smooth reconstruction across the entire European domain. That is the problem that should be challenged by future research.

\subsection{Future Work}
The present study primarily focuses on fields distributed across a geographic coordinate system. An important direction for future research would be to incorporate topographic factors such as elevation, aspect, and slope, as numerous studies have demonstrated their significant influence on spatial and climatic field patterns \cite{diaz2003climate,he2019impact,pepin2022climate}. Including these topographic variables could enhance both the accuracy and the physical relevance of the reconstructed fields.


\section*{Impact statement}
By addressing the critical challenge of data sparsity, our findings demonstrate the effectiveness of the simple inverse distance weighting (IDW) method in reconstructing observational datasets. We believe this study represents a meaningful step toward more efficient and reliable modelling of climate and meteorological systems, paving the way for further advancements in climate science and its practical applications.

Moreover, we show that simple and easily interpretable models can often outperform more advanced approaches due to their computational efficiency and the smaller, more manageable hyper-parameter search space.

\section*{Acknowledgement}
We are grateful to Dr. Stefania A. Ciliberti\href{https://orcid.org/0000-0002-8561-7805}{\includegraphics[width=0.015\textwidth]{orcid.png}} and Dr. Marco Mancini\href{https://orcid.org/0000-0002-9150-943X}{\includegraphics[width=0.015\textwidth]{orcid.png}} for their invaluable advice on methodological issues and remarks regarding the first draft of the paper.

The calculations mentioned in this paper ware performed using the BlueOcean computational cluster which is part of TUL Computing \& Information Services Center infrastructure.


{\appendix
\section*{Supplementary materials}
All scripts, environment setup,  reproducibility containers definitions, and results were published under MIT license and are available as contribution to \textit{Climatrix} project \footnote{The project remote repository is available under the following address: \url{https://github.com/jamesWalczak/climatrix}.}.
}

\bibliography{example_paper}
\bibliographystyle{ieeetr}

\newpage

\begin{IEEEbiography}[{\includegraphics[width=1in,height=1.25in,clip,keepaspectratio]{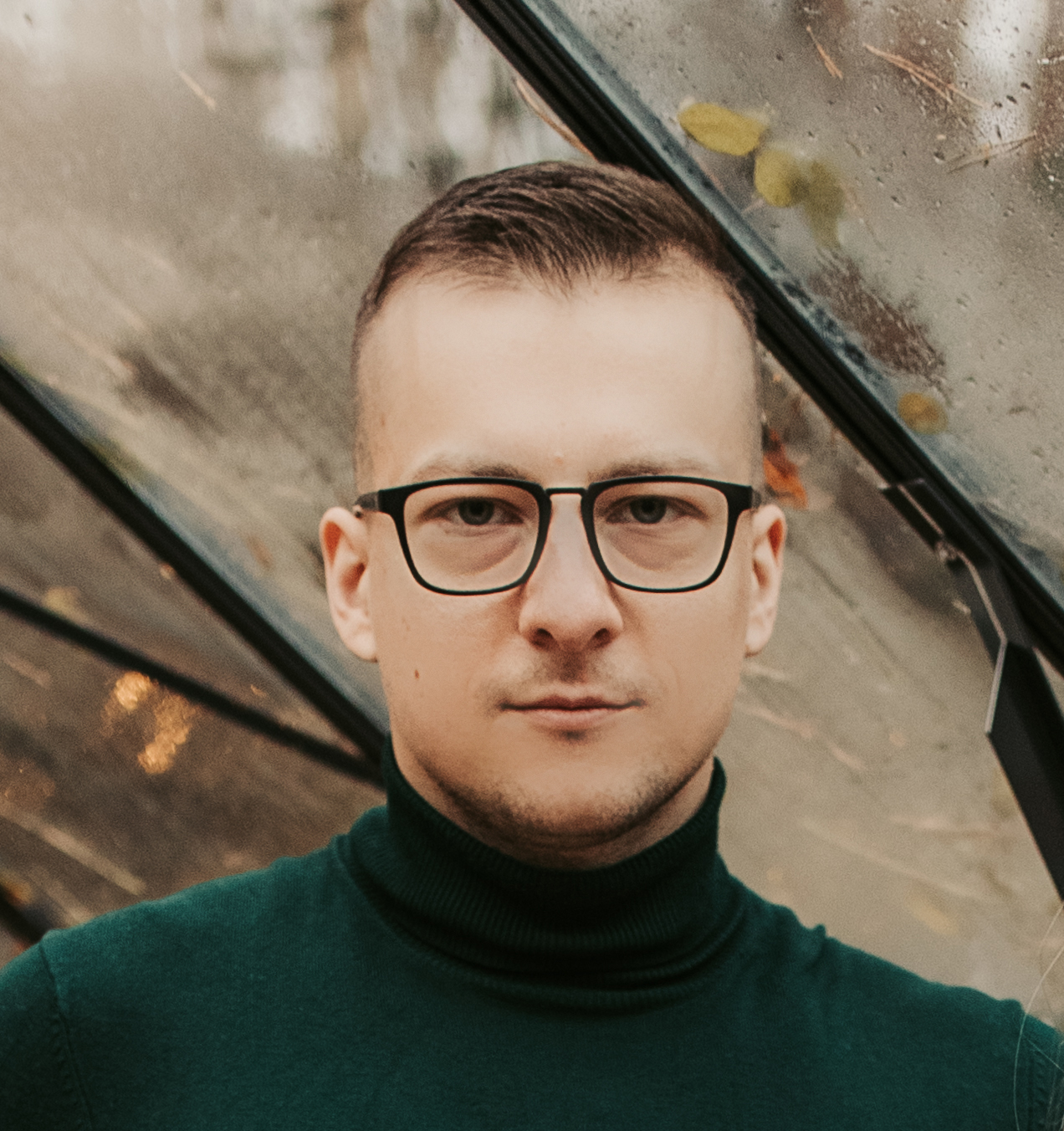}}]{Jakub Walczak}
defended his PhD in 2022, he is an AI researcher with both academic and industrial experience across a wide range of domains, including climate and ICT industry. He specializes in the ethical and multidisciplinary applications of artificial intelligence, bio-inspired advancements, and the development of novel AI paradigms.
\end{IEEEbiography}

\vfill

\end{document}